\pdfoutput=1

\documentclass[11pt]{article}

\usepackage{acl}

\usepackage{times}
\usepackage{latexsym}

\usepackage[T1]{fontenc}

\usepackage[utf8]{inputenc}

\usepackage{microtype}

\usepackage{inconsolata}

\usepackage{multirow}
\usepackage{multicol}
\usepackage{booktabs}
\usepackage{graphicx}
\usepackage{tikz}
\usepackage{paralist}
\usepackage{booktabs}
\usepackage{amsfonts}
\usepackage{CJKutf8}
\usepackage{subfigure}

\usepackage[ruled,vlined]{algorithm2e}
\usepackage{amssymb}
\usepackage{enumitem}
\usepackage{setspace}
\usepackage{amsmath} 
\usepackage{amssymb}
\usepackage{xcolor}

\usepackage{url}
\usepackage{booktabs}
\usepackage{amssymb}
\usepackage{bbding}
\usepackage{pifont}
\usepackage{wasysym}
\usepackage{utfsym}
\usepackage{fontawesome}
\usepackage{footnote} 

\title{Don't Ignore Dual Logic Ability of LLMs while Privatizing: \\ A Data-Intensive Analysis in Medical Domain}

\author{\textbf{
Yanrui Du,
Sendong Zhao\thanks{\llap{}\:\:\:Corresponding author},
Muzhen Cai,
Ming Ma,
Danyang Zhao,
Jiawei Cao,
Bing Qin
}\\
    Harbin Institute of Technology, Harbin, China \\  
    \{ yrdu,sdzhao,mzcai,mma,dyzhao,jwcao,qinb\}@ir.hit.edu.cn\\
}

\begin{document}
\maketitle

\begin{abstract}

Extensive studies have been devoted to privatizing general-domain Large Language Models (LLMs) as Domain-Specific LLMs via feeding specific-domain data. However, these privatization efforts often ignored a critical aspect: Dual Logic Ability, which is a core reasoning ability for LLMs. The dual logic ability of LLMs ensures that they can maintain a consistent stance when confronted with both positive and negative statements about the same fact. Our study focuses on how the dual logic ability of LLMs is affected during the privatization process in the medical domain. We conduct several experiments to analyze the dual logic ability of LLMs by examining the consistency of the stance in responses to paired questions about the same fact. In our experiments, interestingly, we observed a significant decrease in the dual logic ability of existing LLMs after privatization. Besides, our results indicate that incorporating general domain dual logic data into LLMs not only enhances LLMs' dual logic ability but also further improves their accuracy. These findings underscore the importance of prioritizing LLMs' dual logic ability during the privatization process. Our study establishes a benchmark for future research aimed at exploring LLMs' dual logic ability during the privatization process and offers valuable guidance for privatization efforts in real-world applications.

\end{abstract}

\section{Introduction}




Large Language Models (LLMs)\cite{openai2023gpt4,touvron2023llama,baichuan2023baichuan2,du2022glm} have gained widespread attention for their remarkable ability to engage with humans. While LLMs have shown exceptional performance in general domains, their performance in specific domains, such as medicine, remains less established. Consequently, there is a growing interest\cite{wang2023huatuo} in privatizing LLMs to enhance their specific domain expertise.

\begin{figure}[ht]
\centering
\includegraphics[scale=0.45]{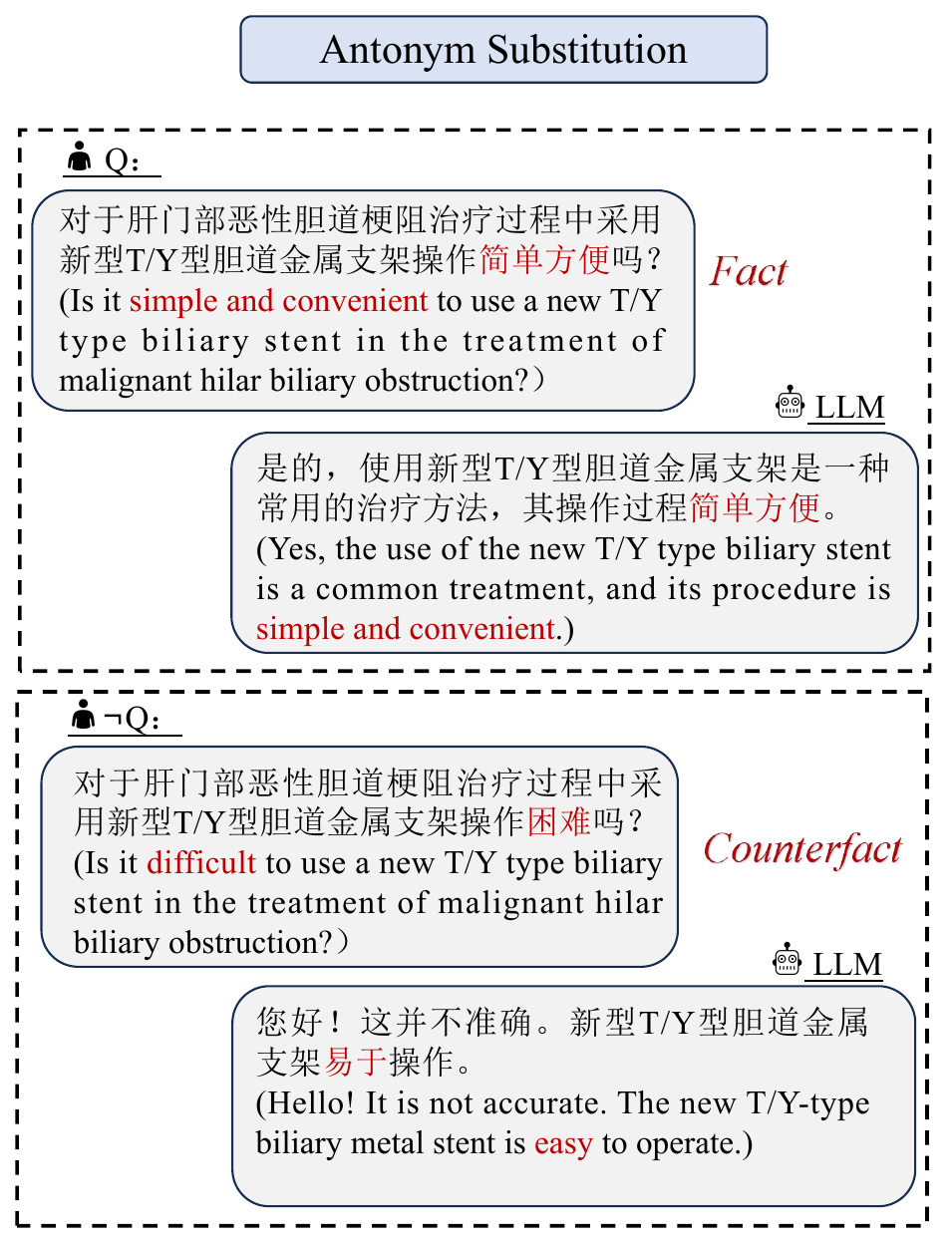}
\caption{An example where LLM maintains a consistent stance when faced with $Q$ and $\neg Q$. Antonym substitution represents the operation of converting $Q$ into $\neg Q$.}
\label{fig:case}
\end{figure}



Recent work\cite{chen2023bianque,Sunsimiao, wang2023huatuo} has aimed to construct high-quality medical domain instruction fine-tuning data based on authoritative medical resources, thereby privatizing LLMs. Despite claims from numerous privatized LLMs that they outperform their base LLMs in medical domains, an ignored aspect is the LLMs' dual logic ability. The concept\footnote{https://en.wikipedia.org/wiki/Duality\_(mathematics)\#} of dual logic has been formally defined in the field of mathematics, as follows: 
\begin{quote}
In logic, functions or relations A and B are considered dual if $A(\neg x)=\neg B(x)$, where $\neg$ is logical negation.
\end{quote}
Building on this concept, we have introduced the notion of the dual logic ability of LLMs, which highlights their ability to maintain a consistent stance when faced with a question $Q$ with the fact statement and a question $\neg Q$ with the corresponding counterfact statement. Formally, it can be defined as \(\neg f(Q) = f(\neg Q)\), where \(f\) symbolizes the function of LLMs. Fig.~\ref{fig:case} provides an illustrative example of how LLMs maintain a consistent stance when confronted with both \(Q\) and \(\neg Q\).

Given the limited focus on the dual logic ability of privatized LLMs in prior research, there exists a notable gap in high-quality data available for analysis. To address this deficiency, we constructed a comprehensive medical dataset. Our work involves collecting literature abstracts and utilizing ChatGPT to automatically generate multi-round dialogues based on these abstracts. Considering that the redundancy of literature content will pose challenges to ChatGPT's precise comprehension, we have developed a multi-step process aimed at enhancing ChatGPT's understanding of content, thereby ensuring the generation of high-quality dialogues. Subsequently, we manually annotate pairs of dual logic test samples \(Q\) and \(\neg Q\) with three specific operations. Furthermore, to enhance the dual logic ability of privatized LLMs we developed an automatic process to construct general domain dual logic data. Overall, our dataset includes medical literature abstracts for the pre-training (PT) stage, medical multi-round dialogues and general domain dual logic data for the instruction fine-tuning (IFT) stage, and medical dual logic test data.

In our experiments, we conducted a data-intensive analysis. Firstly, on our annotated dual logic test data, we evaluate the dual logic ability of existing LLMs and their privatized variants.  Our findings reveal that LLMs' dual logic ability all significantly decreases after privatization, indicating that the dual-logic ability is always ignored in existing efforts. To further explore how LLMs' dual logic ability is affected during the privatization process, we private LLMs on our constructed data. Our experiment results indicate that only using medical multi-round dialogues during the IFT stage often diminishes the LLM's dual logic ability. Conversely, incorporating general domain dual logic data during the IFT stage will significantly improve LLMs' dual logic ability, and their accuracy will further be improved accordingly. These results underscore the importance of LLMs' dual logic ability during the privatization process and demonstrate that LLMs can generalize the ability from general domain dual logic data. Moreover, we observe that the pre-training data can serve as an enhancement factor in improving LLMs' dual logic ability. Through quantitative analyses, we also examined the impact of the base LLM itself and verified the transferability of general domain dual logic data. It is worth noting that we manually evaluated all 10,908 test samples to ensure the reliability of our results.








Our contributions can be summarized as follows:

\begin{itemize}[leftmargin=*,noitemsep,topsep=0pt]
\item Our study introduces the notion of dual logic ability, which is a core reasoning ability for LLMs. Our analysis reveals that LLMs' dual logic ability has been consistently ignored in prior privatization efforts.

\item Our study reveals the importance of LLMs' dual ability during the privatization process and demonstrates that general domain dual logic data plays a key role in improving the dual logic ability of privatized LLMs.

\item Our study provides a comprehensive private dataset that can serve as a benchmark for future work exploring LLMs' dual logic ability during the privatization process.
\end{itemize}

\section{Background}

LLMs have shown remarkable proficiency in the general domain, yet their performance in specific domains reveals considerable scope for enhancement. In the medical domain, when individuals turn to LLMs for medical consultations or to gather health-related information, the depth and professionalism of the LLMs' knowledge becomes paramount. Therefore, numerous studies focused on privatizing LLMs to enhance their expertise. These efforts mainly adopt a uniform technical method, which involves the parameter-efficient fine-tuning of LLMs based on high-quality medical-domain data. For collecting high-quality data, current studies adopt different strategies as follows:
\begin{itemize}[leftmargin=*,noitemsep,topsep=0pt]
\item Med-PaLM\cite{singhal2023large} synthesizes high-quality medical NLP datasets such as PubMedQA\cite{jin2019pubmedqa}, MedMCQA\cite{pal2022medmcqa}, and MedQA~\cite{jin2021disease}. By instruction fine-tuning PaLM, Med-PaLM achieves unprecedented performance in medical multiple-choice tasks.
\item HuatuoGPT\cite{huatuogpt-2023} and ChatDoctor\cite{li2023chatdoctor} focuses on collecting real-world question-answering data and creating medical data via distilling from ChatGPT.
\item Bentsao\cite{wang2023huatuo} and Shennong\cite{zhu2023ChatMed} utilize ChatGPT to construct question-answering data based on the existing structured medical knowledge graph.
\item MedicalGPT-zh\cite{MedicalGPT-zh} referred to BELLE \cite{belle2023exploring} method, utilizing ChatGPT to construct dialogue data in various scenarios based on clinical guideline texts.
\item ChatMed\cite{zhu2023ChatMed} gathers real-world questions from online consultations and then utilizes ChatGPT to respond, thus obtaining question-answering data.
\end{itemize}
While significant efforts are being made to private data in the medical domain, the emphasis is largely on gathering diverse and high-quality medical domain data. Currently, there's a lack of focus on the dual logic ability of privatized LLMs. Therefore, our work offers a more comprehensive privatized dataset specifically designed to explore LLMs' dual logic ability, distinguishing our work from current practices.

\section{Data Construction} \label{data_construct}

\begin{figure*}[ht]
\centering
\includegraphics[scale=0.32]{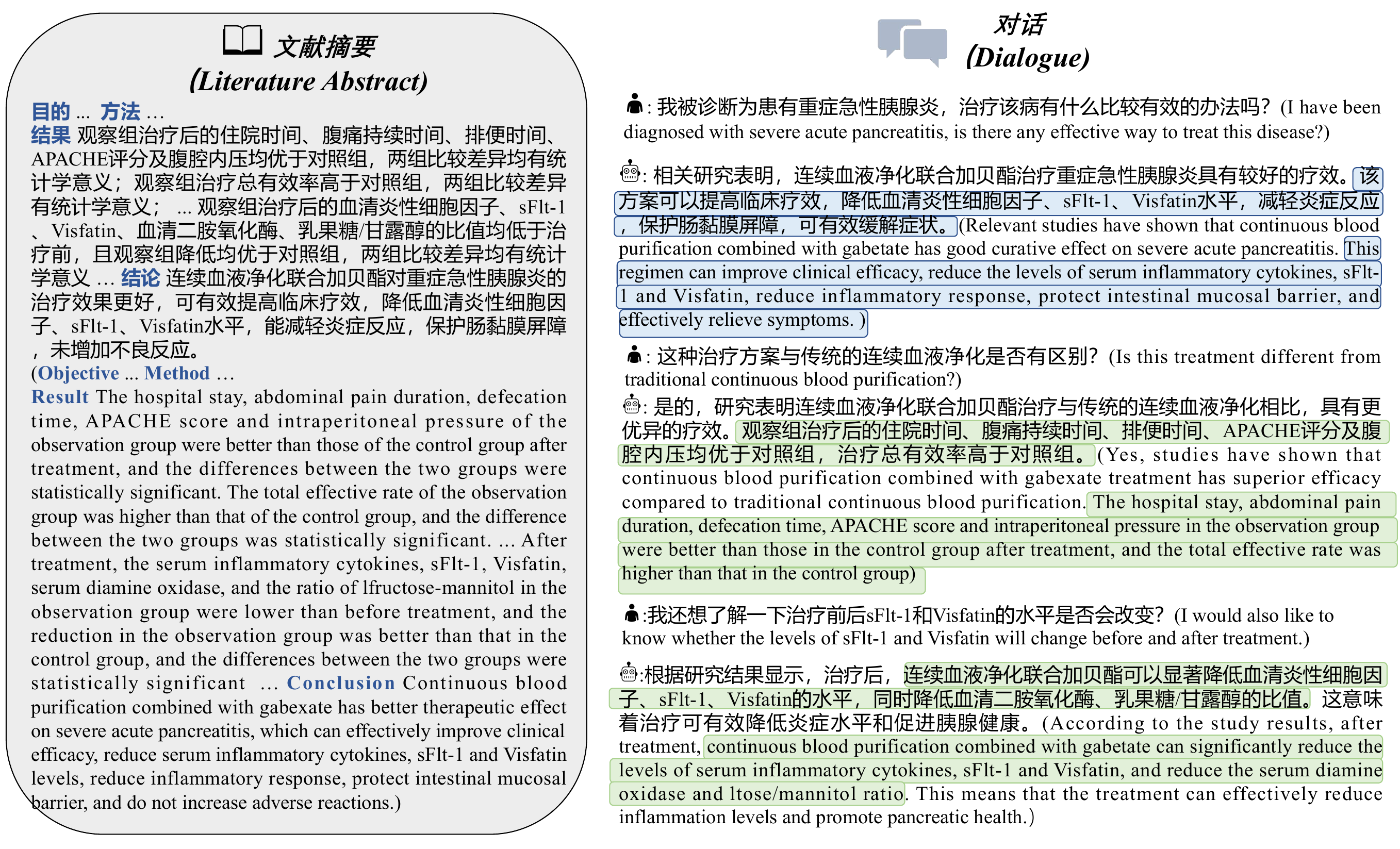}
\caption{A case of constructed dialogue. The left side shows the literature abstract and the right side shows the constructed dialogue. The text covered in green color represents information from the literature abstract.}
\label{fig:dia_case}
\end{figure*}


In our data construction process, we approach it from two perspectives. Firstly, we developed a medical dataset that includes medical literature abstracts, multi-round dialogue automatically generated based on these abstracts, and manually annotated dual logic test data. Secondly, to improve the dual logic ability of LLMs, we systematically constructed dual logic data from the general domain.

\subsection{A Medical Dataset}
\paragraph{Literature abstracts.} We have collected medical literature abstracts from China HowNet~\footnote{https://www.cnki.net/}, focusing on a valuable field ``hepatobiliary and pancreatic diseases''. As shown in Fig.~\ref{fig:dia_case}, these abstracts typically encompass details such as the study's purpose, methods, experiments, and conclusion, offering a comprehensive overview. We collected 10,313 literature abstracts, with an average of 16.22 sentences and 712.21 tokens per abstract.

\paragraph{Multi-round dialogue.}
Dialogue data plays a crucial role in training LLMs during the IFT stage. With an increasing focus on leveraging ChatGPT for creating dialogue data, it's important to ensure ChatGPT's precision comprehension to help construct high-quality dialogue. Therefore, we propose a multi-step process for generating dialogue data based on literature abstracts, divided into the following steps:



\begin{figure*}[ht]
\centering
\includegraphics[scale=0.32]{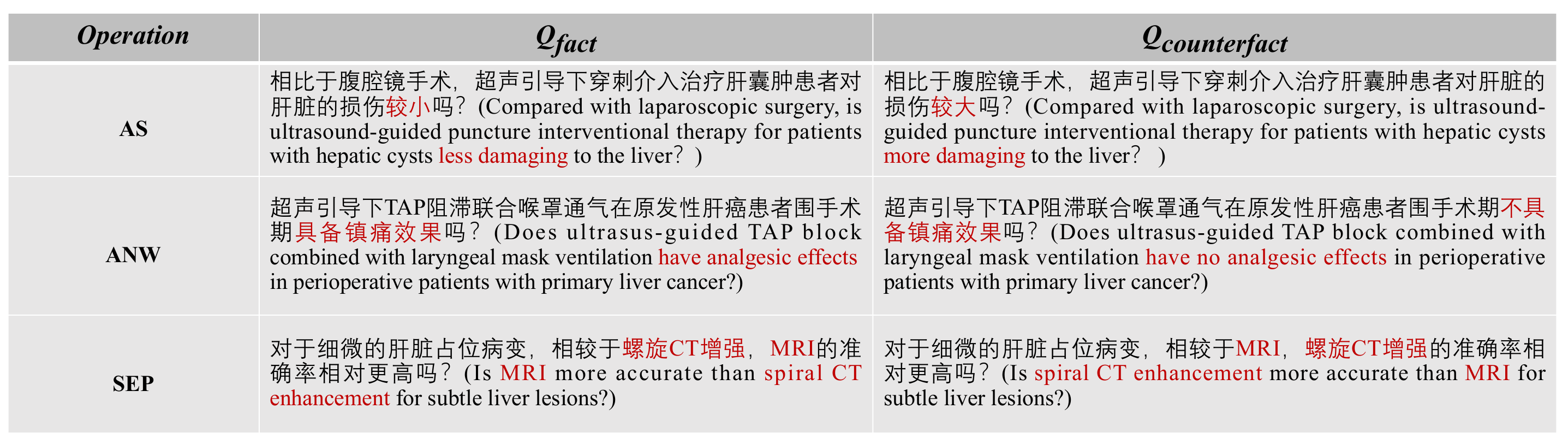}
\caption{Cases of our annotated dual logic test data. AS, ANW, and SEP represent antonym substitution, adding negative words, and swapping entity positions operations respectively.}
\label{fig:dual_test_case}
\end{figure*}

\begin{figure*}[ht]
\centering
\includegraphics[scale=0.45]{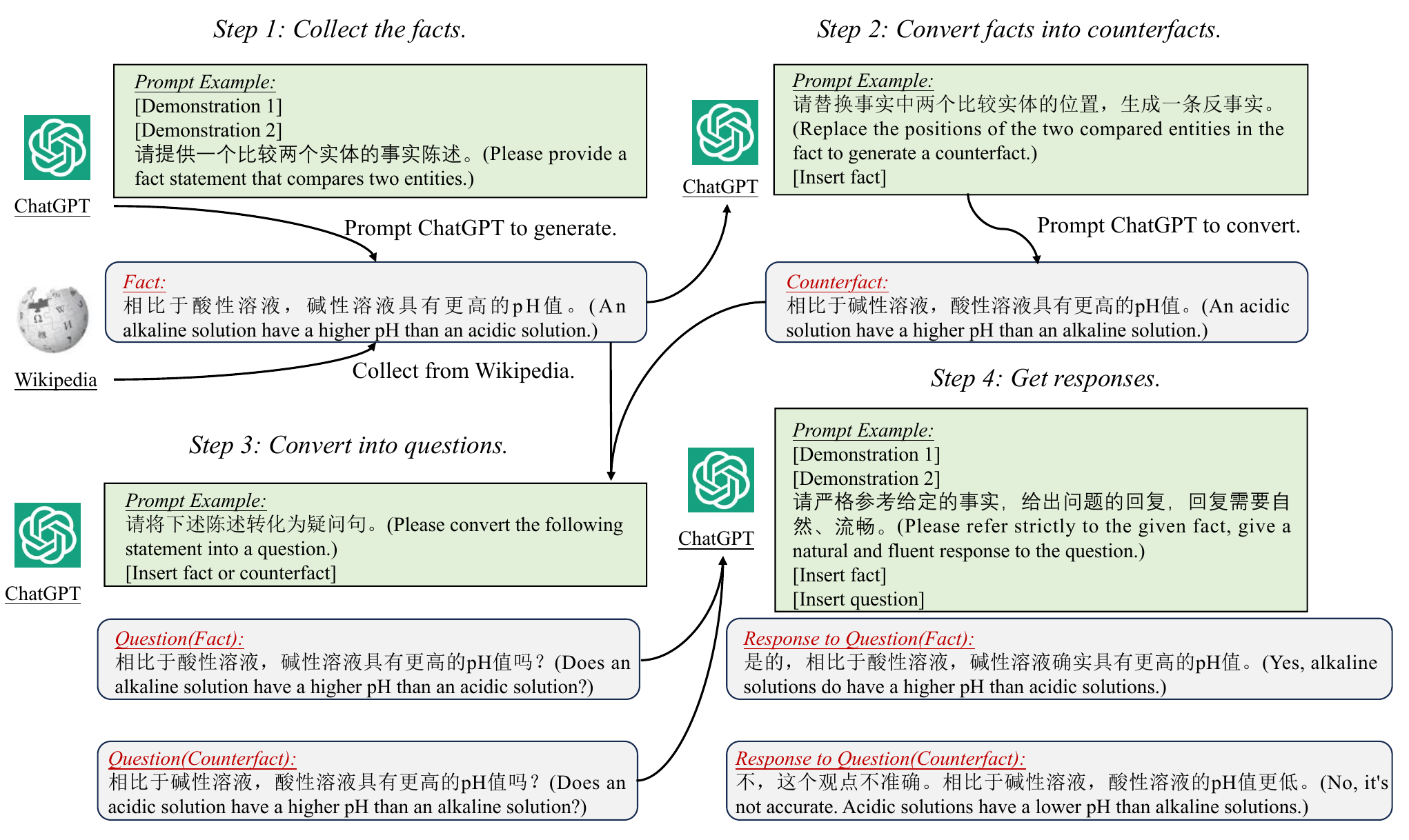}
\caption{Process of constructing general domain dual logic data.}
\label{fig:ad_process}
\end{figure*}

\begin{itemize}[leftmargin=*,noitemsep,topsep=0pt]
\item Information Disassembly: Transform unstructured medical literature abstracts into a semi-structured format. We employ regular expression techniques for key information extraction, including the study's purpose, methods, experiments, and conclusions. When regular expression techniques fail, ChatGPT can be further engaged for information extraction.
\item Prompt Focus: Recent studies\cite{shi2023large} suggest that LLMs always become distracted when processing excessively lengthy text. To mitigate this issue, prompts can be strategically designed to direct the LLMs' attention towards specific information that needs to be incorporated into the dialogue. For instance, the prompt can be designed as ``Please incorporate the [conclusion] into the dialogue. Other details, such as the [purpose], [method], [results], etc., can be referenced as needed.''.
\item Reverse Verification: Empirical observations indicate that generated dialogue always incorporates the original text content from medical literature. Consequently, we employ the longest common substring matching algorithm to assess the extent of literature information incorporation, allowing us to selectively preserve high-quality dialogue.
\item Data Cleaning: Generated dialogues frequently encompass a substantial of subjective expressions, exemplified by phrases like, ``This study shows that…''. When such data are utilized in the training stage, LLMs naturally replicate similar subjective expressions. During interactions with humans, it will lead to the emergence of perplexing subjective statements. To mitigate this issue, we employ ChatGPT to rephrase such subjective expressions into objective expressions. For instance, transforming ``This study shows that ...'' into ``Relevant study indicates that...''.
\end{itemize}

We constructed 1,212 multi-round dialogues with an average of 3.53 rounds per dialogue. As shown in Fig.~\ref{fig:dia_case}, following the above steps, we constructed a three-round dialogue data, in which the information of the first round is supplemented by ChatGPT, and the information of the second and third rounds comes from literature abstracts.

\paragraph{Dual logic test data.}

To evaluate the dual logic ability of LLMs, we meticulously annotated dual logic test data. This process entailed providing annotators with only literature abstracts, ensuring that our constructed dialogues remained closed to prevent the risk of training data leakage. The annotators were first tasked with distilling fact statements from abstracts. Subsequently, they transformed these fact statements into counterfact statements through a specific operation (antonyms substitution, adding negative words, or swapping entity positions). Finally, annotators formulate both fact statements and their corresponding counterfact statements into general questions to obtain pairs of dual logic test data. Overall, we annotated 100 pairs for antonyms substitution operation, 58 pairs for adding negative words operation, and 44 pairs for swapping entity positions operation. Fig.~\ref{fig:dual_test_case} illustrates cases of our annotated dual logic test data.






\subsection{General Domain Dual Logic Data}


To enhance the dual logic ability of LLMs, we employ ChatGPT to automate the generation of dual logic data from the general domains. The process is outlined in Fig.~\ref{fig:ad_process}. Initially, we gathered the fact statements in two ways: sourcing from Wikipedia and generating via prompting ChatGPT. Subsequently, we utilize ChatGPT to generate the counterfact statements via operations such as antonyms substitution, adding negation words, and swapping entity positions. Following this step, we prompt ChatGPT to transform both fact and counterfact statements into their corresponding general questions. Finally, we instruct ChatGPT to craft responses that are based on the original fact statement. Overall, we construct 130 pairs for antonyms substitution operation, 150 pairs for adding negation words operation, and 100 pairs for swapping entity positions operation.

\section{Experiment}

\begin{figure}[t]
\centering
\includegraphics[scale=0.35]{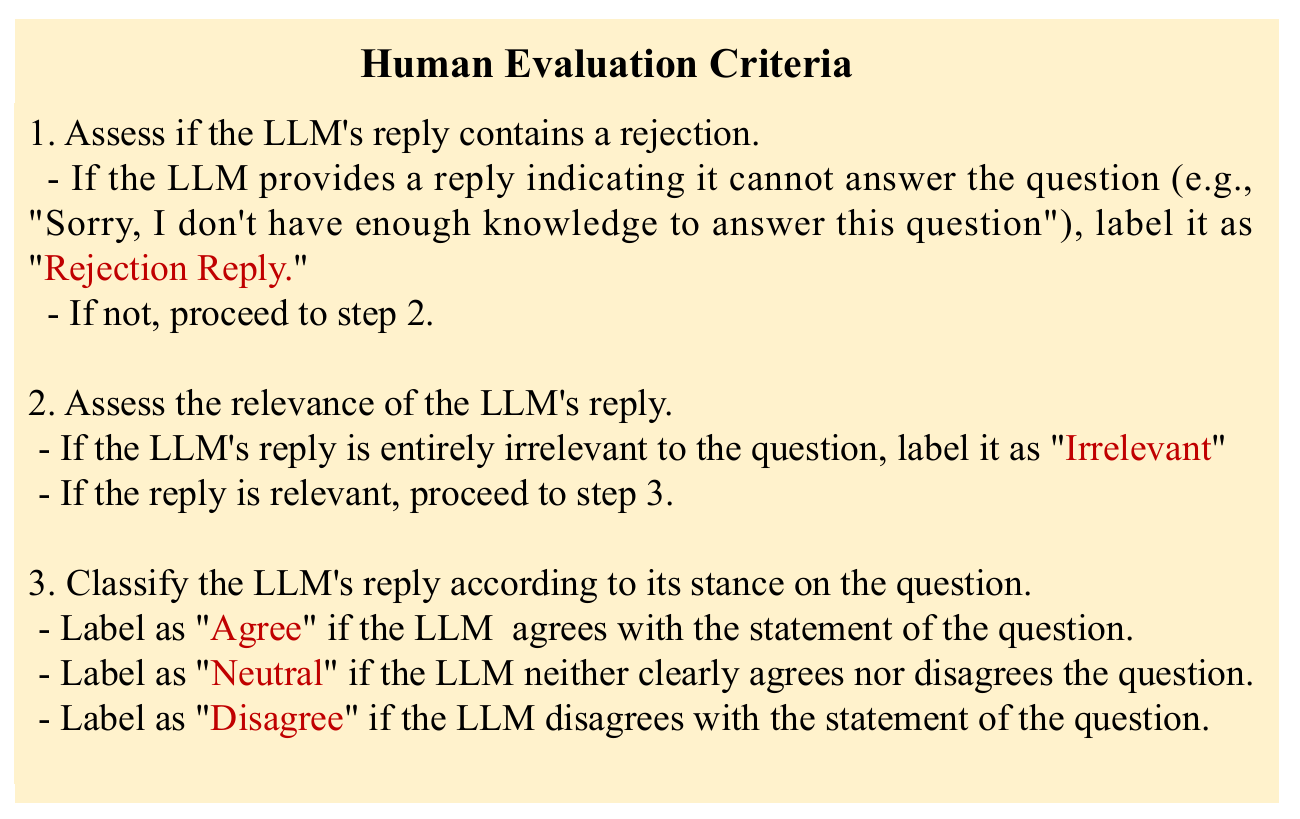}
\caption{Huamn evaluation criteria. In our evaluation, LLMs' replies will be classified into five categories.}
\label{fig:huamn_eval}
\end{figure}

\subsection{Preliminary}

Before presenting the experiment results, we introduce our evaluation metrics, selected LLMs, and experiment settings in our experiments.

\paragraph{Evaluation metrics.} 

In our study, we manually annotate all evaluation results, according to the criteria illustrated in Fig.~\ref{fig:huamn_eval}. In evaluating dual logic ability, we regard a question with the fact statement and a question with the corresponding counterfact statement as a paired test sample. The paired test samples are counted if they satisfy the following conditions: 
\begin{itemize}[leftmargin=*,noitemsep,topsep=0pt]
\item For the question with the fact statement, the LLM maintains an agreement stance while for the question with the corresponding counterfact statement, the LLM maintains a disagreement stance.
\item For both the question with the fact statement and the question with the counterfact statement, the LLM maintains a neutral stance.
\item For the question with the fact statement, the LLM maintains a disagreement stance while for the question with the corresponding counterfact statement, the LLM maintains an agreement stance.
\end{itemize}
By calculating the proportion of paired samples that adhered to the above conditions out of the total number, we calculate the \textbf{Dual-Logic Index (DL-Index)}.

Accuracy is another evaluation metric in our study. Notably, unlike the dual-logic index, accuracy is calculated based on individual test samples rather than pairs. Specifically, we calculate the proportion of the following test samples:
\begin{itemize}[leftmargin=*,noitemsep,topsep=0pt]
\item For the question with the fact statement, the LLM maintains an agreement stance.
\item For the question with the counterfact statement, the LLM maintains a disagreement stance.
\end{itemize}

Besides, to simultaneously evaluate dual logic ability and accuracy, we draw inspiration from the F1 index commonly used in machine learning. DL-Acc F1 will be calculated as:

\begin{gather}
F1 = 2 \times \frac{\text{DL-Index} \times \text{Accuracy}}{\text{DL-Index} + \text{Accuracy}}
\end{gather}






\begin{table}[t]
\begin{tabular}{l|ccc}
\toprule[0.7pt]
    & Qwen$_{7B}$   & Huozi2$_{7B}$  & ChatGLM2$_{6B}$ \\
\midrule[0.5pt]
PT  & 8.27\% & 14.59\% & 4.32\%   \\
IFT & 0.07\% & 0.05\%  & 0.06\%  \\
\bottomrule[0.7pt]
\end{tabular}
\caption{Percentage of LLMs' updated parameters. PT and IFT represent the pre-training stage and the instruction fine-tuning stage respectively. }
\label{tab:size_pa}
\end{table}

\paragraph{Experiment settings.}
In our study, on the one hand, we utilize our annotated dual logic test data to evaluate the dual logic ability of existing LLMs. There has been substantial progress in transforming general-domain LLMs into medical-domain LLMs based on their respective private medical data. Among them, we have selected the CareQwen$_{14B}$\cite{wang2023caregpt} built upon Qwen$_{14B}$\cite{bai2023qwen}, the Bentsao$_{7B}$\cite{wang2023huatuo} built upon Huozi$_{7B}$\footnote{https://github.com/HIT-SCIR/huozi}, and the Bianque2$_{6B}$\cite{chen2023bianque} built upon ChatGLM$_{6B}$\cite{du2022glm} as our analysis objects.

On the other hand, considering that the training process of existing LLMs is black-box, it is difficult to perform a fair and comprehensive analysis. To address this issue, we conduct the transformation of general-domain LLMs into medical-domain LLMs based on our private data. During this process, we assess the influence of different types of data on the LLMs' performance. We have selected Qwen$_{7B}$, Huozi2$_{7B}$, and ChatGLM2$_{6B}$ as the foundational LLMs for training. During the training process, we have adapted the parameter-efficient fine-tuning LoRA\cite{peft,hu2021lora} strategy. Tab.~\ref{tab:size_pa} illustrates the percentage of LLMs' updated parameters, including both the pre-training (PT) stage and the instruction fine-tuning (IFT) stage. Experience with the size of updated parameters comes from the open-source study~\footnote{https://github.com/ymcui/Chinese-LLaMA-Alpaca}.


\subsection{Main Experiment} \label{sec_main_exp}

\begin{figure}[t]
\centering
\subfigure[Analysis on Qwen$_{14B}$ and CareQwen$_{14B}$.]{
\centering
\includegraphics[scale=0.45]{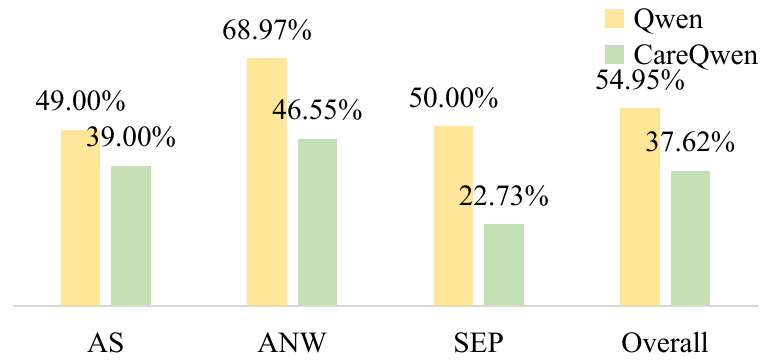}
}
\subfigure[Analysis on Huozi$_{7B}$ and Bentsao$_{7B}$.]{
\centering
\includegraphics[scale=0.45]{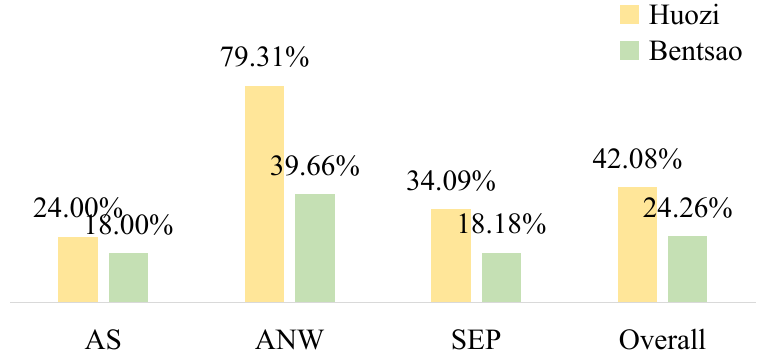}
}
\subfigure[Analysis on Bianque2$_{6B}$ and ChatGLM$_{6B}$.]{
\centering
\includegraphics[scale=0.45]{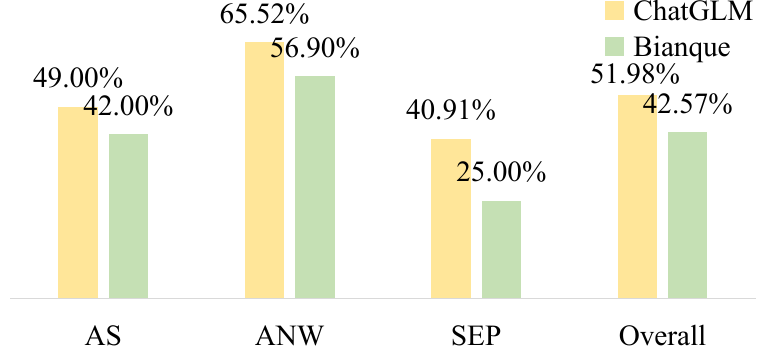}
}
\caption{Evaluate the dual logic ability of existing LLMs. We report the DL-Index across three operations (AS, ANW, and SEP) as well as an overall assessment.}
\label{fig:ana_existing_llms}
\end{figure}

\paragraph{Evaluation on existing LLMs.}

As shown in Fig.~\ref{fig:ana_existing_llms}, we separately conduct evaluations of the dual logic ability for both the foundational LLMs and their corresponding privatized variants. The experimental results suggest that under each fine-grained operation and overall, training LLMs with their privatized data leads to a stable and significant decrease in the dual logic ability. This phenomenon underscores a prevalent oversight in current work, where the dual logic ability of LLMs during the privatization process is ignored.

\begin{table*}[t]
\setlength{\tabcolsep}{3.5pt}
\centering
\begin{tabular}{l|cccc|cccc|c}
\toprule[0.7pt]
\multicolumn{1}{l|}{}         & \multicolumn{4}{c|}{DL-Index}                                                   & \multicolumn{4}{c|}{Accuracy}                                                              & F1                   \\
                             & AS                   & ANW                  & SEP                  & Overall              & AS                   & ANW                  & SEP                  & Overall              &                      \\
\midrule[0.5pt]
Qwen$_{7B}$                         &                      &                      &                      &                      &                      &                      &                      &                      &                      \\
Base                         & 37.00\%              & 65.52\%              & 18.18\%              & 41.09\%              & 61.00\%              & 68.97\%              & 48.86\%              & 60.64\%              & 48.99\%              \\
IFT$^{\spadesuit}$                          & 49.00\%              & 36.21\%              & 22.73\%              & 39.60\%              & 72.50\%              & 64.66\%              & 61.36\%              & 67.82\%              & 50.01\%              \\
IFT$^{\clubsuit}$                        & 64.00\%              & 79.31\%              & 40.91\%              & 63.37\%              & 79.00\%              & 84.48\%              & 61.36\%              & 76.73\%              & 69.41\%              \\
PT\&IFT$^{\spadesuit}$                      & 66.00\%              & 68.97\%              & 36.36\%              & 60.40\%              & 80.00\%              & 75.86\%              & 60.23\%              & 74.50\%              & 66.71\%              \\
PT\&IFT$^{\clubsuit}$                   & \textbf{72.00\%}              & \textbf{86.21\%}              & \textbf{47.73\%}              &  \textbf{70.79\%}              & \textbf{81.00\%}              & \textbf{89.66\%}              & \textbf{70.45\%}              & \textbf{81.19\%}              & \textbf{75.63\%}              \\
\midrule[0.5pt]
\multicolumn{1}{l|}{Huozi2$_{7B}$} & \multicolumn{1}{l}{} & \multicolumn{1}{l}{} & \multicolumn{1}{l}{} & \multicolumn{1}{l}{} & \multicolumn{1}{l}{} & \multicolumn{1}{l}{} & \multicolumn{1}{l}{} & \multicolumn{1}{l}{} & \multicolumn{1}{l}{} \\
Base                         & 36.00\%              & \textbf{79.31\%}              & \textbf{36.36\%}              & 48.51\%              & 57.50\%              & 75.00\%              & 52.27\%              & 61.39\%              & 54.20\%              \\
IFT$^{\spadesuit}$                          & 29.00\%              & 29.31\%              & 13.64\%              & 25.74\%              & 62.00\%              & 68.97\%              & 54.55\%              & 62.38\%              & 36.14\%              \\
IFT$^{\clubsuit}$                       & 53.00\%              & 70.69\%              & 20.45\%              & 50.99\%              & \textbf{74.50\%}              & 80.17\%              & 57.95\%              & 72.52\%              & 59.88\%              \\
PT\&IFT$^{\spadesuit}$                      & 27.00\%              & 56.90\%              & 15.91\%              & 33.17\%              & 61.50\%              & 75.00\%              & 55.68\%              & 64.11\%              & 43.72\%              \\
PT\&IFT$^{\clubsuit}$                    & \textbf{60.00\%}              & 70.69\%             & 25.00\%              & \textbf{55.45\%}              & 74.00\%              & \textbf{83.62\%}              & \textbf{61.36\%}              & \textbf{74.01\%}              & \textbf{63.40\%}              \\
\midrule[0.5pt]
\multicolumn{1}{l|}{ChatGLM2$_{6B}$} & \multicolumn{1}{l}{} & \multicolumn{1}{l}{} & \multicolumn{1}{l}{} & \multicolumn{1}{l}{} & \multicolumn{1}{l}{} & \multicolumn{1}{l}{} & \multicolumn{1}{l}{} & \multicolumn{1}{l}{} & \multicolumn{1}{l}{} \\
Base                         & 44.00\%              & 67.24\%              & 20.45\%              & 45.54\%              & 52.50\%              & 67.24\%              & 45.45\%              & 55.20\%              & 49.91\%              \\
IFT$^{\spadesuit}$                           & 63.00\%              & 68.97\%              & 11.36\%              & 53.47\%              & 77.50\%              & 79.31\%              & 53.41\%              & 72.77\%              & 61.64\%              \\
IFT$^{\clubsuit}$                        & 62.00\%              & \textbf{79.31\%}              & 29.55\%              & 59.90\%              & 74.00\%              & 86.21\%              & 62.50\%              & 75.00\%              & 66.61\%              \\
PT\&IFT$^{\spadesuit}$                       & 55.00\%              & 55.17\%              & 18.18\%              & 47.03\%              & 74.50\%              & 75.86\%              & 54.55\%              & 70.54\%              & 56.44\%              \\
PT\&IFT$^{\clubsuit}$                   & \textbf{78.00\%}              & \textbf{79.31\%}              & \textbf{31.82\%}              & \textbf{68.32\%}              & \textbf{85.50\%}              & \textbf{82.76\%}              & \textbf{63.64\%}              & \textbf{79.95\%}              & \textbf{73.68\%}       \\
\bottomrule[0.7pt]
\end{tabular}
\caption{Experimental results on our private data. PT and IFT represent the pre-training stage and the instruction fine-tuning stage respectively. Compared to ${\spadesuit}$, ${\clubsuit}$ represents the mixing of general domain dual logic data.}
\label{main_exp}
\end{table*}

\paragraph{Evaluation on our LLMs.}
As introduced in Sec.~\ref{data_construct}, our private dataset comprises medical literature abstracts, multi-round medical dialogues, and general domain dual logic data. Building on this foundation, we implemented four settings in the training process of our LLMs:

\begin{itemize}[leftmargin=*,noitemsep,topsep=0pt]
\item \textbf{IFT$^{\spadesuit}$}: IFT conducted only based on multi-round medical dialogues.
\item \textbf{IFT$^{\clubsuit}$}: IFT conducted based on both multi-round medical dialogues and general domain dual logic data.
\item \textbf{PT\&IFT$^{\spadesuit}$}: Following pre-trainining on medical literature abstracts, perfrom the setting IFT$^{\spadesuit}$.
\item \textbf{PT\&IFT$^{\clubsuit}$}: Following pre-trainining on medical literature abstracts, perfrom the setting IFT$^{\clubsuit}$.

\end{itemize}
According to the experimental results shown in Tab.~\ref{main_exp}, we observed the following phenomena: 
\begin{itemize}[leftmargin=*,noitemsep,topsep=0pt]

\item The accuracy can not reflect the dual logic capacity of LLMs. For instance, in the IFT$^{\spadesuit}$ setting, where despite Qwen and ChatGLM demonstrating improved accuracy, their dual logic ability experienced a decline. This observation underscores the necessity for evaluations of LLMs' dual logic ability.


\item In the IFT$^{\spadesuit}$ setting, relying solely on multi-round dialogue data during the IFT stage typically leads to a decline in LLMs' dual logic ability. Conversely, in the IFT$^{\clubsuit}$ setting, incorporating general domain dual logic data into the IFT stage not only significantly improves LLMs' dual logic ability levels but also further improves their accuracy. These findings suggest that LLMs can effectively generalize dual logic ability from general domain dual logic data and highlight the importance of LLMs' dual logic ability.



\item Pre-training (PT) data serves as an enhancement rather than a critical factor in improving dual logic ability. Results based on Huozi2 indicate that although the dual logic ability under the PT\&IFT$^{\spadesuit}$ setting shows improvement over the IFT$^{\spadesuit}$ setting, it does not even surpass that of base LLM itself. Furthermore, on ChatGLM2, the PT\&IFT$^{\spadesuit}$ setting even shows a decrease in performance compared to the IFT$^{\spadesuit}$ setting. These observations indicate that only incorporating the PT stage can not stably yield improvements in dual logic ability. However, when the PT stage is incorporated under the IFT$^{\clubsuit}$ setting, we observe a stable and significant improvement (the PT\&IFT$^{\clubsuit}$ setting) across all three LLMs. This phenomenon indicates that general domain dual logic data plays a critical factor while pre-training data can serve as an enhancing factor.


\end{itemize}

\begin{figure}[t]
\centering
\includegraphics[scale=0.45]{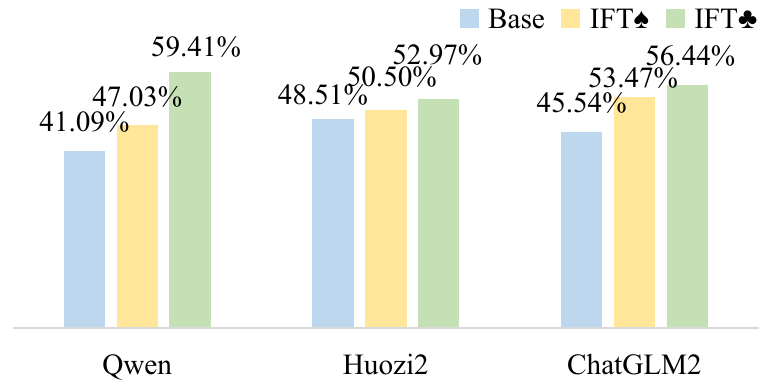}
\caption{Experimental results of mixing general domain dual logic data into another private dataset.}
\label{fig:ad_other_data}
\end{figure}

\begin{figure}[ht]
\centering
\subfigure[Analysis on Qwem$_{7B}$.]{
\centering
\includegraphics[scale=0.45]{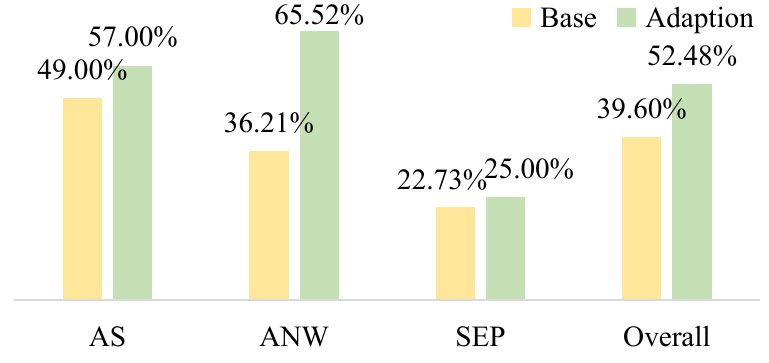}
}
\subfigure[Analysis on Huozi2$_{7B}$.]{
\centering
\includegraphics[scale=0.45]{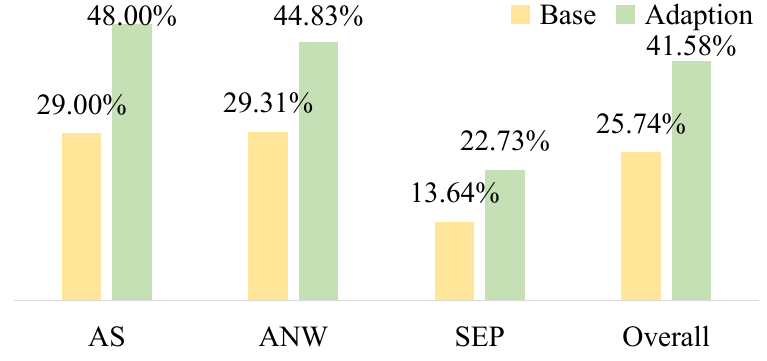}
}
\subfigure[Analysis on ChatGLM2$_{6B}$.]{
\centering
\includegraphics[scale=0.45]{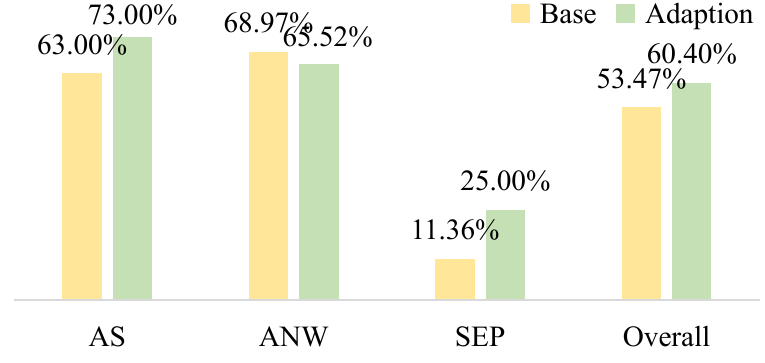}
}
\caption{Analyze the impact of base LLMs. Compared with Base, Adaption represents that the dual logic ability of the base LLM itself has been improved.}
\label{fig:base_effect}
\end{figure}

Overall, our evaluation first underscores the importance of assessing LLMs' dual logic ability. Then, by conducting a detailed analysis of our private data, we observed the influence of different types of data on dual logic ability. Our findings indicate that relying solely on medical domain dialogue often compromises LLMs' dual logic ability. In the area of improving LLMs' dual logic ability during privatizing, general domain dual logic can serve as a critical factor while pre-training data can serve as an enhancing factor.




\section{Exploration}

\subsection{Transferability of general domain dual logic data}
We explore whether our constructed general domain dual logic data can be transferred to other privatized data\footnote{https://github.com/SCIR-HI/Huatuo-Llama-Med-Chinese}, rather than being specific to our privatized data. To verify this, we collect another open-source private dataset. Under the IFT$^{\clubsuit}$ setting as introduced in Sec.~\ref{sec_main_exp}, we train three LLMs. The experimental results as shown in Fig.~\ref{fig:ad_other_data} suggest that after mixing our constructed general domain dual logic data, the dual logic ability of LLMs is significantly improved. This phenomenon indicates that our constructed general domain dual logic data can serve as a valuable resource for transferring to any private data.

\subsection{Impact of base LLMs}
We conduct a quantitative experiment to analyze the impact of the base LLM itself on dual logic ability. According to Sec.~\ref{sec_main_exp}, we have concluded that LLMs can generalize to dual logic ability from general domain dual logic data. Therefore, we conduct an experimental setup (called \textbf{Adaption}), that is, first train LLMs on general domain dual logic data to improve their dual logic ability, and then train on privatized data to transform them into medical domain LLMs. The results, illustrated in Fig.~\ref{fig:base_effect}, demonstrate a significant improvement in the dual logic ability of the privatized LLMs following the enhancement of the base LLMs' dual logic ability. This phenomenon indicates that it is important and valuable to improve the dual logic ability of base LLM itself.





\subsection{Case Study}
As shown in App,~\ref{app:case_study}, we provide a real case study based on Huozi2. Under the base and IFT$^{\spadesuit}$ settings, LLMs both fail to maintain a consistent stance when faced with $Q$ and $\neg Q$. In contrast, under the IFT$^{\clubsuit}$ setting which incorporates general domain dual logic data, LLM can maintain a consistent stance, that it agrees with $Q$ and disagrees with $\neg Q$.

\section{Conclusion and Future Work}

In our research, we introduce the notion of dual logic ability of LLMs. Our analysis highlights the critical need to evaluate their dual logic ability during the privatization process, an aspect frequently ignored in prior studies. Through our data-intensive analysis within the medical domain, we demonstrate that LLMs can generalize the dual logic ability from general domain dual logic data. Moreover, our constructed dataset can serve as a benchmark for future research on exploring LLMs' dual logic ability during the privatization process, and our work offers valuable guidance for privatization efforts in real-world applications. In future work, we suggest that based on our datasets, further studies can explore the influence of other factors, such as training strategy and the space of parameters, on LLMs' dual logic ability during the privatization process.

\section{Limitation}
\begin{itemize}[leftmargin=*,noitemsep,topsep=0pt]
\item Pre-training corpus for our medical dataset is small-scale, which is influenced by several factors. Firstly, due to permission restrictions, we are unable to access the full text of literature from HowNet, thus limiting us to using only literature abstracts for our training corpus. Besides, the privatization process always aims for high efficiency and resource conservation, making the construction of a large-scale pre-training corpus counterproductive to these goals. However, it's important to acknowledge that expanding the size of the pre-training corpus may yield different experimental results.

\item In our exploration of LLMs' dual logic ability, we have applied only three operations and focused exclusively on the task form of stance detection. This raises two important considerations for future research. Firstly, it is possible to incorporate more operations to construct dual logic data. Secondly, it is valuable to explore the construction of dual logic data based on the form of other tasks.
\end{itemize}

\section{Ethics Statement}

For our work, we ensure that our constructed dataset is devoid of any content that could compromise personal privacy or cause harm. Our team of annotators is not only professionally trained but also adheres to a strict disclaimer to uphold ACL standards. Besides, in our work, we utilized the ChatGPT API interface from November 15, 2023 to November 30, 2023. Our experiments are conducted on open-source LLMs, for which we secured explicit authorization from the authors responsible for their maintenance.


\bibliography{anthology,custom}

\begin{thebibliography}{21}
\expandafter\ifx\csname natexlab\endcsname\relax\def\natexlab#1{#1}\fi

\bibitem[{Bai et~al.(2023)Bai, Bai, Chu, Cui, Dang, Deng, Fan, Ge, Han, Huang et~al.}]{bai2023qwen}
Jinze Bai, Shuai Bai, Yunfei Chu, Zeyu Cui, Kai Dang, Xiaodong Deng, Yang Fan, Wenbin Ge, Yu~Han, Fei Huang, et~al. 2023.
\newblock Qwen technical report.
\newblock \emph{arXiv preprint arXiv:2309.16609}.

\bibitem[{Baichuan(2023)}]{baichuan2023baichuan2}
Baichuan. 2023.
\newblock \href {https://arxiv.org/abs/2309.10305} {Baichuan 2: Open large-scale language models}.
\newblock \emph{arXiv preprint arXiv:2309.10305}.

\bibitem[{Chen et~al.(2023)Chen, Wang, Xing, huimin zheng, Xu, Fang, Wang, Li, Wu, Liu, and Xu}]{chen2023bianque}
Yirong Chen, Zhenyu Wang, Xiaofen Xing, huimin zheng, Zhipei Xu, Kai Fang, Junhong Wang, Sihang Li, Jieling Wu, Qi~Liu, and Xiangmin Xu. 2023.
\newblock \href {http://arxiv.org/abs/2310.15896} {Bianque: Balancing the questioning and suggestion ability of health llms with multi-turn health conversations polished by chatgpt}.

\bibitem[{Du et~al.(2022)Du, Qian, Liu, Ding, Qiu, Yang, and Tang}]{du2022glm}
Zhengxiao Du, Yujie Qian, Xiao Liu, Ming Ding, Jiezhong Qiu, Zhilin Yang, and Jie Tang. 2022.
\newblock Glm: General language model pretraining with autoregressive blank infilling.
\newblock In \emph{Proceedings of the 60th Annual Meeting of the Association for Computational Linguistics (Volume 1: Long Papers)}, pages 320--335.

\bibitem[{Hu et~al.(2021)Hu, Shen, Wallis, Allen-Zhu, Li, Wang, Wang, and Chen}]{hu2021lora}
Edward~J Hu, Yelong Shen, Phillip Wallis, Zeyuan Allen-Zhu, Yuanzhi Li, Shean Wang, Lu~Wang, and Weizhu Chen. 2021.
\newblock Lora: Low-rank adaptation of large language models.
\newblock \emph{arXiv preprint arXiv:2106.09685}.

\bibitem[{Ji et~al.(2023)Ji, Deng, Gong, Peng, Niu, Zhang, Ma, and Li}]{belle2023exploring}
Yunjie Ji, Yong Deng, Yan Gong, Yiping Peng, Qiang Niu, Lei Zhang, Baochang Ma, and Xiangang Li. 2023.
\newblock Exploring the impact of instruction data scaling on large language models: An empirical study on real-world use cases.
\newblock \emph{arXiv preprint arXiv:2303.14742}.

\bibitem[{Jin et~al.(2021)Jin, Pan, Oufattole, Weng, Fang, and Szolovits}]{jin2021disease}
Di~Jin, Eileen Pan, Nassim Oufattole, Wei-Hung Weng, Hanyi Fang, and Peter Szolovits. 2021.
\newblock What disease does this patient have? a large-scale open domain question answering dataset from medical exams.
\newblock \emph{Applied Sciences}, 11(14):6421.

\bibitem[{Jin et~al.(2019)Jin, Dhingra, Liu, Cohen, and Lu}]{jin2019pubmedqa}
Qiao Jin, Bhuwan Dhingra, Zhengping Liu, William~W Cohen, and Xinghua Lu. 2019.
\newblock Pubmedqa: A dataset for biomedical research question answering.
\newblock \emph{arXiv preprint arXiv:1909.06146}.

\bibitem[{Li et~al.(2023)Li, Li, Zhang, Dan, Jiang, and Zhang}]{li2023chatdoctor}
Yunxiang Li, Zihan Li, Kai Zhang, Ruilong Dan, Steve Jiang, and You Zhang. 2023.
\newblock Chatdoctor: A medical chat model fine-tuned on a large language model meta-ai (llama) using medical domain knowledge.
\newblock \emph{Cureus}, 15(6).

\bibitem[{Liu et~al.(2023)Liu, Liao, Meng, Wang, and Wang}]{MedicalGPT-zh}
Hongcheng Liu, Yusheng Liao, Yutong Meng, Yu~Wang, and Yanfeng Wang. 2023.
\newblock \url{https://github.com/MediaBrain-SJTU/MedicalGPT-zh}.

\bibitem[{Mangrulkar et~al.(2022)Mangrulkar, Gugger, Debut, Belkada, and Paul}]{peft}
Sourab Mangrulkar, Sylvain Gugger, Lysandre Debut, Younes Belkada, and Sayak Paul. 2022.
\newblock Peft: State-of-the-art parameter-efficient fine-tuning methods.
\newblock \url{https://github.com/huggingface/peft}.

\bibitem[{Ming~Wang(2023)}]{Sunsimiao}
Dong~Xue* Ming~Wang, Xin~Yan. 2023.
\newblock Sunsimiao: Chinese medical large language model.
\newblock \url{https://github.com/X-D-Lab/Sunsimiao}.

\bibitem[{OpenAI(2023)}]{openai2023gpt4}
OpenAI. 2023.
\newblock \href {http://arxiv.org/abs/2303.08774} {Gpt-4 technical report}.

\bibitem[{Pal et~al.(2022)Pal, Umapathi, and Sankarasubbu}]{pal2022medmcqa}
Ankit Pal, Logesh~Kumar Umapathi, and Malaikannan Sankarasubbu. 2022.
\newblock Medmcqa: A large-scale multi-subject multi-choice dataset for medical domain question answering.
\newblock In \emph{Conference on Health, Inference, and Learning}, pages 248--260. PMLR.

\bibitem[{Shi et~al.(2023)Shi, Chen, Misra, Scales, Dohan, Chi, Sch{\"a}rli, and Zhou}]{shi2023large}
Freda Shi, Xinyun Chen, Kanishka Misra, Nathan Scales, David Dohan, Ed~H Chi, Nathanael Sch{\"a}rli, and Denny Zhou. 2023.
\newblock Large language models can be easily distracted by irrelevant context.
\newblock In \emph{International Conference on Machine Learning}, pages 31210--31227. PMLR.

\bibitem[{Singhal et~al.(2023)Singhal, Azizi, Tu, Mahdavi, Wei, Chung, Scales, Tanwani, Cole-Lewis, Pfohl et~al.}]{singhal2023large}
Karan Singhal, Shekoofeh Azizi, Tao Tu, S~Sara Mahdavi, Jason Wei, Hyung~Won Chung, Nathan Scales, Ajay Tanwani, Heather Cole-Lewis, Stephen Pfohl, et~al. 2023.
\newblock Large language models encode clinical knowledge.
\newblock \emph{Nature}, 620(7972):172--180.

\bibitem[{Touvron et~al.(2023)Touvron, Lavril, Izacard, Martinet, Lachaux, Lacroix, Rozi{\`e}re, Goyal, Hambro, Azhar et~al.}]{touvron2023llama}
Hugo Touvron, Thibaut Lavril, Gautier Izacard, Xavier Martinet, Marie-Anne Lachaux, Timoth{\'e}e Lacroix, Baptiste Rozi{\`e}re, Naman Goyal, Eric Hambro, Faisal Azhar, et~al. 2023.
\newblock Llama: Open and efficient foundation language models.
\newblock \emph{arXiv preprint arXiv:2302.13971}.

\bibitem[{Wang et~al.(2023{\natexlab{a}})Wang, Liu, Xi, Qiang, Zhao, Qin, and Liu}]{wang2023huatuo}
Haochun Wang, Chi Liu, Nuwa Xi, Zewen Qiang, Sendong Zhao, Bing Qin, and Ting Liu. 2023{\natexlab{a}}.
\newblock Huatuo: Tuning llama model with chinese medical knowledge.
\newblock \emph{arXiv preprint arXiv:2304.06975}.

\bibitem[{Wang et~al.(2023{\natexlab{b}})Wang, Zhou, Chen, Wang, and Tan}]{wang2023caregpt}
Rongsheng Wang, Ruizhe Zhou, Haoming Chen, Yapeng Wang, and Tao Tan. 2023{\natexlab{b}}.
\newblock Caregpt: Medical llm, open source driven for a healthy future.
\newblock \url{https://github.com/WangRongsheng/CareGPT}.

\bibitem[{Zhang et~al.(2023)Zhang, Chen, Jiang, Yu, Chen, Li, Chen, Wu, Zhang, Xiao, Wan, Wang, and Li}]{huatuogpt-2023}
Hongbo Zhang, Junying Chen, Feng Jiang, Fei Yu, Zhihong Chen, Jianquan Li, Guiming Chen, Xiangbo Wu, Zhiyi Zhang, Qingying Xiao, Xiang Wan, Benyou Wang, and Haizhou Li. 2023.
\newblock Huatuogpt, towards taming language models to be a doctor.
\newblock \emph{arXiv preprint arXiv:2305.15075}.

\bibitem[{Zhu and Wang(2023)}]{zhu2023ChatMed}
Wei Zhu and Xiaoling Wang. 2023.
\newblock Chatmed: A chinese medical large language model.
\newblock \url{https://github.com/michael-wzhu/ChatMed}.

\end{thebibliography}
\appendix

\section{Case Study}\label{app:case_study}

\begin{figure*}[hb]
\centering
\includegraphics[scale=0.35]{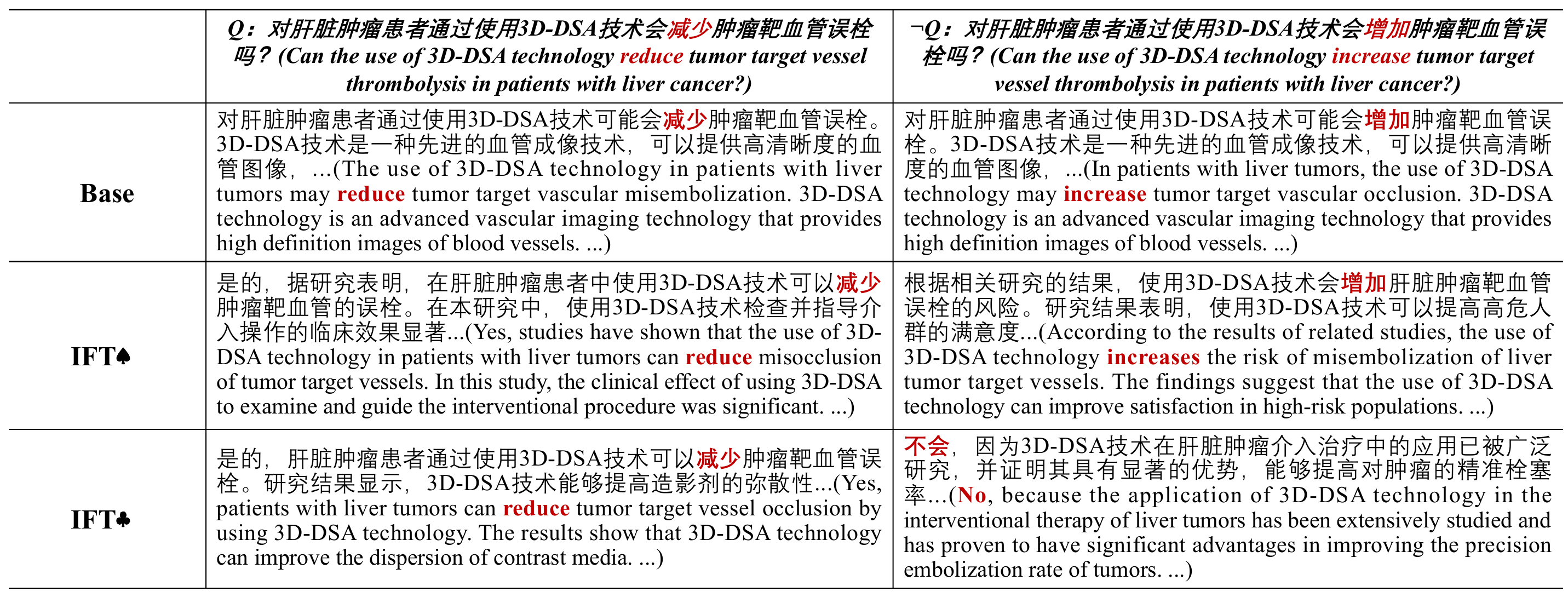}
\caption{A real case study based on Huozi2.}
\label{fig:case_study}
\end{figure*}

As shown in Fig.~\ref{fig:case_study}, we present a real case study based on Huozi2 to illustrate the improvement effect of general domain dual logic data on private LLMs' dual logic ability.

\end{document}